\pdfoutput=1
\documentclass[pmlr,twocolumn,10pt]{jmlr}

\providecommand{\mlhtrack}[1]{}
\mlhtrack{findings}

    \jmlrproceedings{}{ML4H 2026 - Findings Track}
    \jmlrworkshop{Machine Learning for Health (ML4H) 2026}

\usepackage{booktabs}
\usepackage{array}
\usepackage{graphicx}
\usepackage[T1]{fontenc}
\usepackage{tikz}
\usetikzlibrary{positioning,arrows.meta}
\title[Observed, Curated, Reused]{Observed, Curated, Reused: Observation Bias in Clinical AI Evaluation}
\author[Ahmed et al.]{\Name{Md Shamim Ahmed} \Email{shamim@imada.sdu.dk}\\
 \Name{Lukas {Galke Poech}} \Email{galke@imada.sdu.dk}\\
 \Name{Richard R\"ottger} \Email{roettger@imada.sdu.dk}\\
 \addr Department of Mathematics and Computer Science (IMADA), University of Southern Denmark, Odense, Denmark}

\begin{document}
\maketitle

\begin{abstract}
Clinical AI evaluations are vulnerable to \textbf{observation bias} embedded in routine medical documentation, which captures only what clinicians notice, test, and record rather than a complete clinical picture. As this evidence travels into publications and reference knowledge graphs, crucial distinctions disappear: unexamined symptoms blur with confirmed negatives, and chronological disease progression is lost.
Surviving evidence can also enter both the benchmark query and the reference knowledge base, creating a self-referential loop where a patient is evaluated against clues derived from their own published case report.

To quantify this bias, we tracked 6,503 clinical cases across six historical Human Phenotype Ontology (HPO) releases (October 2021--June 2026) to identify when HPO first cited the publications describing them. This yielded a confirmatory cohort with an observable ``source ingestion'' event, where HPO cited a case's source publication as target-disease evidence. Diagnostic rankers depended heavily on these source publications: for our primary semantic ranker (Resnik), publication-weighted mean reciprocal rank rose from 0.26 before paper ingestion to 0.84 upon inclusion in HPO, but collapsed to 0.05 when source-linked assertions were counterfactually removed. Two additional rankers (IC-cosine and LIRICAL) reproduced this trajectory, and control experiments confirmed the diagnostic jump was driven by the patient's publication rather than background HPO growth.

Clinical AI systems may therefore score well because their knowledge base contains
evidence linked to the test case's source publication, rather than through
generalization. Modern knowledge is not the problem; evaluation evidence
must match the claim. Trustworthy benchmarks must separate patients,
source-linked evidence, and time: historical tests should use only knowledge available at the intended decision time.

\end{abstract}

\begin{keywords}
clinical machine learning; benchmark leakage; observation bias; Human Phenotype Ontology; evidence provenance
\end{keywords}

\paragraph*{Data and Code Availability}
MIMIC-IV and eICU are restricted by their data-use agreements
\citep{johnson2023mimiciv,pollard2018eicu}; HPO and Phenopacket Store are public
\citep{gargano2024hpo,danis2025phenopacketstore}. Code, frozen protocols, configurations,
checksums, and aggregate result artifacts are available at
\url{https://gitlab.sdu.dk/screen4care/observed-curated-reuse}; restricted records are not shared.

\paragraph*{Institutional Review Board (IRB)}
This study used secondary, de-identified records and public resources without recruitment,
intervention, or re-identification; formal research-ethics approval was not required.

\begin{figure*}[!t]
    \raggedleft
    \includegraphics[width=1\textwidth]
    {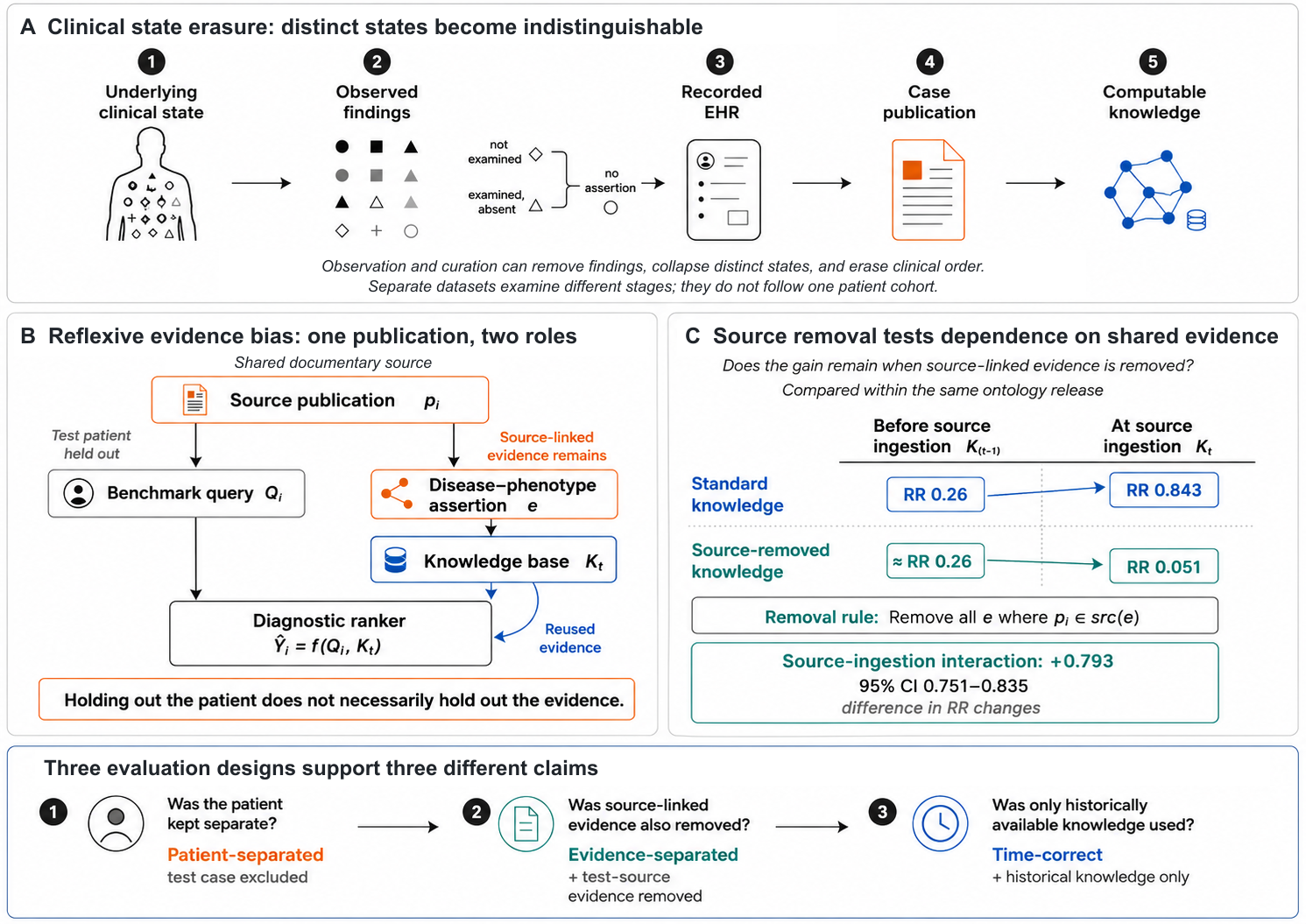}
    \caption{\textbf{Evidence loss and reuse.} \textbf{A}, Curation and observation erase clinical distinctions (state erasure). \textbf{B}, Reusing a publication for benchmark queries and knowledge creates reflexive evidence bias. \textbf{C}, Testing performance before and after ingestion, with and without source annotations, isolates source dependence under three evaluation boundaries.}
    \label{fig:design-findings}
\end{figure*}

\section{Introduction}

Clinical evidence is fundamentally shaped by what clinicians notice, test, document, publish, and curate \citep{agniel2018biases,sisk2021informative,chang2022disparate}. This selective reporting introduces \textbf{observation bias}, where diagnostic models risk learning the behavioral patterns of healthcare systems rather than true disease pathology. Furthermore, as evidence moves from bedside charts into publications and structured knowledge bases, critical context disappears: unexamined findings can become indistinguishable from confirmed negatives, and coarse timestamps erase recorded clinical order. We term this informational flattening \textbf{clinical state erasure}.

Literature-derived cases are increasingly used to benchmark diagnostic rankers and language models \citep{danis2025phenopacketstore,bridges2025pheval,reese2026systematic}. The rankers we study are not trained on patients in the usual machine-learning sense. They take a patient's symptoms, written as HPO terms \citep{gargano2024hpo}, compare them with the symptom profile that HPO keeps for each disease, and return the diseases in order of how well they match. Everything such a ranker knows comes from these profiles, and each entry in a profile cites its source, often a published case report. In these evaluations, a single publication can supply both the patient features used to construct a test query and the disease--phenotype annotations used to rank diagnoses. Consider a case report describing a child with a rare disorder. A benchmark turns the child's recorded symptoms into a test query; when HPO curators later process the same report, they add those symptoms to the disorder's profile. Thus, although the patient is nominally held out, evidence from their case report may remain available to the system. We call this shared-source dependency \textbf{reflexive evidence bias}.

The three problems arise at different points along the same path. Observation bias decides what is written down in the first place, state erasure decides how much of it survives curation, and reflexive evidence bias decides whether what survives is then used to grade the very patient it came from.

We test whether diagnostic performance artificially jumps once HPO incorporates the case report describing the patient. Holding patient symptoms constant, we evaluate diagnostic rankings before and after each paper entered the ontology, both with and without its specific evidence. Control tests confirm that general ontology growth cannot explain these gains, pipeline audits trace where clinical details are lost to state erasure, and mitigation experiments evaluate whether scoring adjustments can restore reliable diagnoses (Figure~\ref{fig:design-findings}).

\section{Study Design}

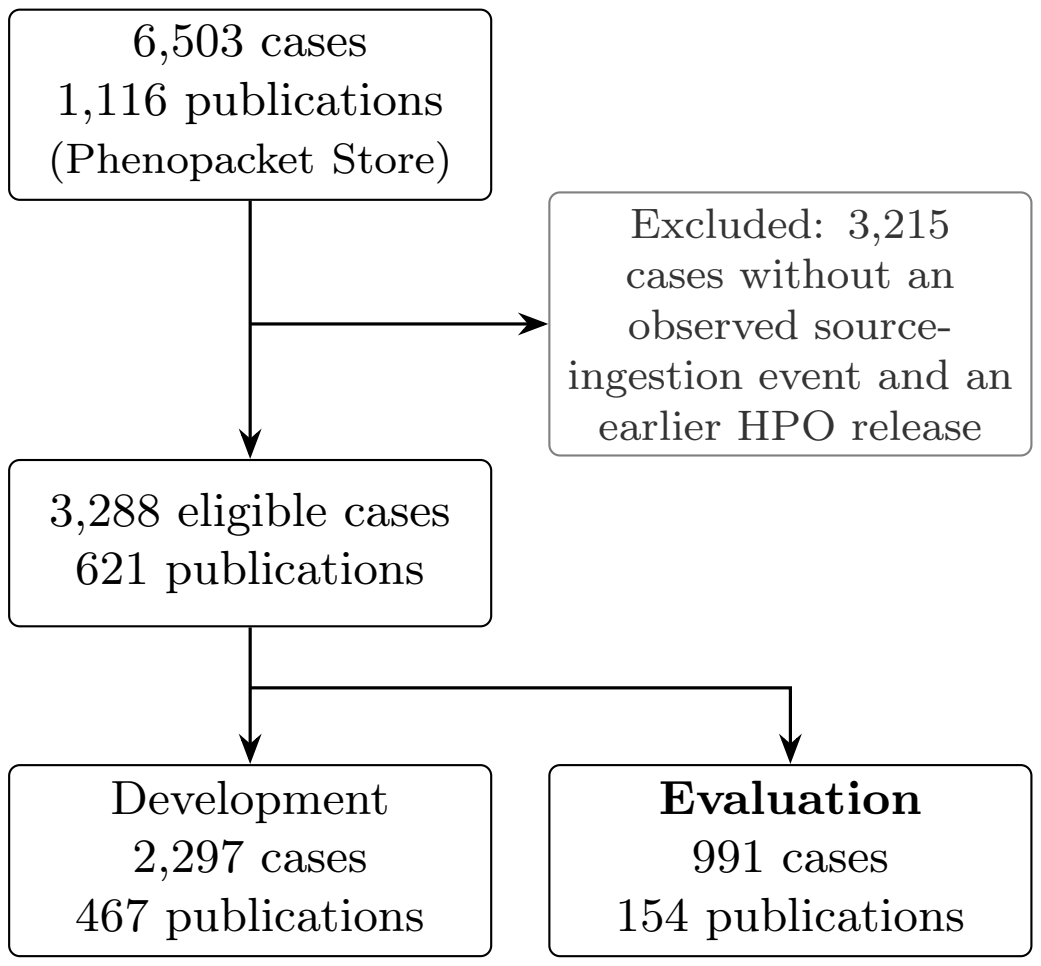
\begin{figure}[t]
    \centering
    \small
    \begin{tikzpicture}[
        box/.style={draw, rounded corners=2pt, align=center, inner sep=3pt,
            text width=0.40\columnwidth, minimum height=11mm,
            execute at begin node={\hyphenpenalty=10000}},
        excl/.style={box, draw=gray, text=black!80, font=\footnotesize},
        arr/.style={-{Stealth[length=2mm]}, semithick}]
        \node[box] (all) at (-0.24\columnwidth, 0) {6,503 cases\\1,116 publications\\{\footnotesize(Phenopacket Store)}};
        \node[excl] (ex) at (0.24\columnwidth, -1.45) {Excluded: 3,215 cases without an observed source-ingestion event and an earlier HPO release};
        \node[box] (elig) at (-0.24\columnwidth, -2.9) {3,288 eligible cases\\621 publications};
        \node[box] (dev) at (-0.24\columnwidth, -5.0) {Development\\2,297 cases\\467 publications};
        \node[box] (test) at (0.24\columnwidth, -5.0) {\textbf{Evaluation}\\991 cases\\154 publications};
        \draw[arr] (all) -- (elig);
        \draw[arr] (all.south |- ex.west) -- (ex.west);
        \draw[arr] (elig) -- (dev);
        \draw[arr] (elig.south) ++(0,-4mm) -| (test.north);
    \end{tikzpicture}
    \caption{\textbf{Cohort selection.} Eligible cases were split by publication (stratified by ingestion release) before any evaluation, holding out 25\% of publications (154 of 621), which corresponds to 30\% of eligible cases (991 of 3,288). No publication contributes to both partitions.}
    \label{fig:cohort}
\end{figure}

\paragraph{\hyperref[find:source-ingestion]{Does a case become easier when HPO begins citing its source?}}
\label{design:source-ingestion}
We analyzed 6,503 patient cases across six HPO releases between October 2021 and June 2026. Each case is a patient described in a published case report and curated in Phenopacket Store \citep{danis2025phenopacketstore}. The query is simply that patient's recorded phenotypes, as HPO terms; we never build it from HPO's disease profiles, and it stays the same in every condition of the analysis. Evaluating a case requires two snapshots: the release immediately before HPO first cited its case report (\textit{baseline}) and the release where it was first included (\textit{ingestion}). A total of 3,288 cases from 621 publications met this criterion (Figure~\ref{fig:cohort}). Before running any evaluations, we split these cases by publication, so that no publication contributes to both development and evaluation, and held out a random 25\% of publications, stratified by ingestion release. This left an evaluation cohort of 991 cases from 154 publications. Because publications differ in how many patients they describe, 25\% of publications corresponds to 30\% of eligible cases.

We evaluate each case across a $2 \times 2$ design: two timepoints (\textit{baseline} vs. \textit{ingestion}) crossed with two knowledge states (\textit{standard} vs. \textit{source-removed}). For a patient case $Q_i$ from publication $p_i$, the source-removed knowledge base excludes every assertion citing that paper:
\begin{equation}
K_t^{-p_i} = \{e \in K_t : p_i \notin \operatorname{src}(e)\},
\end{equation}
where $\operatorname{src}(e)$ lists the publications supporting assertion $e$. Within each release, the patient's query symptoms and HPO hierarchy remain fixed; only disease profiles and term weights are recalculated.

To measure how much a case's diagnostic rank improved while accounting for routine database updates, we compute a difference-in-differences score on Reciprocal Rank ($RR$):
\begin{equation}
\Delta_i = (RR_{iS+} - RR_{iS-}) - (RR_{iR+} - RR_{iR-})
\end{equation}

Here, $S$ (standard) and $R$ (paper-removed) track knowledge before ($-$) and after ($+$) the paper was added. A positive $\Delta_i$ shows the paper drove the score jump, not normal database updates.
Reciprocal rank ($RR$) is 1.0 for rank 1, scales down for lower ranks, and assigns 0 to failures. To prevent large studies from skewing results, scores are averaged within each publication, weighting all papers equally.

The primary analysis uses directional \hyperref[app:resnik]{Resnik similarity}
\citep{resnik1995information}; we repeat it in the same cohort with
\hyperref[app:cosine]{IC-cosine} and \hyperref[app:lirical]{LIRICAL 2.4.1}
\citep{robinson2020lirical}. Resnik is a long-standing baseline for phenotype-driven diagnosis \citep{kohler2009phenomizer}, and we chose it mainly because it is transparent: each patient term is matched to the most informative ancestor it shares with a disease profile, so we can follow, term by term, how source-linked annotations move a disease up the ranking. Because candidate disease pools differ, we compare within-ranker performance shifts ($\Delta_i$) rather than raw cross-model scores.

\paragraph{\hyperref[find:growth-controls]{Ruling out background HPO growth.}}
\label{design:growth-controls}
To verify that background HPO changes did not cause the performance jump, we tested four alternative factors: adding unrelated articles from the same time period, changing the HPO hierarchy, deleting evidence from competing diseases, and artificially inflating phenotype counts. We also repeated the analysis after removing duplicate cases and after excluding the largest data batch.

\paragraph{\hyperref[find:data-loss]{Auditing upstream data loss.}}
\label{design:data-loss}
Because rankers operate on curated evidence, we separately measured what disappears before models see it. In hospital records (MIMIC-IV and eICU; \citealp{johnson2023mimiciv,pollard2018eicu}), removing encounters or rounding timestamps reduced the recoverability of recorded order across 3,160 diagnosis pairs. Across the medical literature, we tracked 9,722 published disease--phenotype links and measured how many were retained in HPO, PrimeKG \citep{chandak2023primekg}, and OpenBioLink \citep{breit2020openbiolink}.
These tests measure what knowledge resources actually retain, not curator mistakes or true biological disease onset.

\paragraph{\hyperref[find:remedies]{Testing algorithmic remedies.}}
\label{design:remedies}
We tested three approaches to handling incomplete phenotype information and, separately, whether observation-aware scoring could recover performance after source removal.
In total, we evaluated four targeted adjustments: (1) giving more weight to frequent symptoms (752 RareBench cases; \citealp{chen2024rarebench}); (2) penalizing diagnoses when expected symptoms are missing (407 Phenopackets); (3) training neural networks on symptom details such as severity and timing; and (4) adjusting scores for clinician testing patterns (230 benchmark cases).

\section{Findings}

\paragraph{\hyperref[design:source-ingestion]{Source-linked evidence drives diagnostic rankings.}}
\label{find:source-ingestion}
In our primary benchmark, Mean Reciprocal Rank (MRR) rose from 0.26 to 0.84 when the source publication was added to the knowledge base, but collapsed to 0.05 when it was removed. This marks a net gain of +0.793 (95\% CI: 0.751--0.835) driven by the publication alone. Across 991 test cases, 974 benefited directly from including their source paper, none were harmed, and 17 tied. Alternative rankers (IC-cosine and LIRICAL) confirmed this result (Table~\ref{tab:findings-results}).

\paragraph{Why does removal fall below the pre-ingestion score?}
If deleting the paper's evidence simply rewound HPO, performance should return to 0.26. Instead it drops to 0.05, because the two numbers come from different versions of HPO. The 0.26 was measured on the release before the paper was cited. The 0.84 and 0.05 were measured on the same later release, with the same patient queries, and differ only in whether assertions citing the paper are kept. In the meantime HPO changed in other ways: competing diseases gained annotations and term weights were recomputed, and removing one paper undoes none of that. Our removal is also strict. It deletes any assertion that cites the paper, even when another publication supports the same phenotype, so the correct disease is left with less evidence than it would otherwise have. The gap persists when the candidate diseases and the HPO hierarchy are held fixed (0.05 versus 0.26), which points to these annotation changes rather than to newly added diseases. This is why we do not compare scores across releases directly: the difference-in-differences estimate asks, within each release, how much the paper's evidence changes performance.

\paragraph{\hyperref[design:growth-controls]{Background HPO growth does not explain the gain.}}
\label{find:growth-controls}
Adding a patient's source paper to HPO increased diagnostic rank by +0.582, whereas adding unrelated papers from the same time period produced a gain of just +0.002. This +0.580 net difference (95\% CI: 0.521--0.638) supports the conclusion that background HPO growth does not explain the improvement. Control tests further confirmed specificity: deleting clues for the true disease caused diagnostic performance to collapse, while deleting clues for competing diseases had virtually no effect. Real clinical evidence also substantially outperformed adding fake symptoms, indicating that evidence content, rather than added phenotype count alone, mattered (Table~\ref{tab:findings-results}). The improvement also persisted after removing duplicate cases or excluding the largest data batch. Finally, the gain persisted when restricting analysis to cases where the paper was the sole cited source for a symptom (Appendix~\ref{app:control-results}).

\begin{table}[t]
    \caption{\textbf{Source-ingestion effects.} Estimates include 95\% CIs;
    subtraction rows are control contrasts. Exclusive-source analysis uses
    868 cases from 123 publications.}
    \label{tab:findings-results}
    \centering
    \small
    \begin{tabular*}{\columnwidth}{@{\extracolsep{\fill}}>{\raggedright\arraybackslash}p{0.43\columnwidth}rr@{}}
        \toprule
        Analysis & Estimate & 95\% CI \\
        \midrule
        Resnik
            & +0.793 & 0.751--0.835 \\
        IC-cosine
            & +0.427 & 0.364--0.492 \\
        LIRICAL
            & +0.797 & 0.752--0.840 \\
        Event $-$ control
            & +0.580 & 0.521--0.638 \\
        Target removal
            & +0.792 & 0.750--0.833 \\
        Competitor removal
            & +0.001 & 0.000--0.003 \\
        Actual $-$ matched
            & +0.772 & 0.728--0.815 \\
        Exclusive source
            & +0.372 & 0.303--0.442 \\
        \bottomrule
    \end{tabular*}
\end{table}

\paragraph{\hyperref[design:data-loss]{Audits reveal substantial upstream clinical data loss.}}
\label{find:data-loss}
In MIMIC-IV, retaining only 25\% of encounters left both conditions detectable in 79.4\% of diagnosis pairs, but the ability to determine which occurred first collapsed from 76.8\% to 12.9\%. By contrast, removing entire patients at the same total data volume preserved both conditions in 91.9\% of pairs, demonstrating that fragmenting a patient's timeline specifically destroys chronological order. Coarsening timestamps to weekly intervals turned 83.3\% of MIMIC-IV and 90.0\% of eICU diagnosis-pair sequences into uninformative ties. Across literature curation, only 35.5\% (3,454 of 9,722) of published disease--phenotype links were captured in HPO 2021. Crucially, these audits measure documentation recoverability rather than curator error (Appendix~\ref{app:evidence-loss-results}).

\paragraph{\hyperref[design:remedies]{Targeted scoring adjustments failed to recover performance.}}
\label{find:remedies}
Supporting experiments found mixed benefits from phenotype frequency, negative findings, and qualifiers.
Frequency-aware scoring offered no meaningful gain over Resnik (+0.006 MRR; 95\% CI: $-0.008$--$0.019$). For temporally unseen cases, calibrating negative-evidence penalties raised MRR from 0.074 to 0.229, yet simply ignoring negative evidence was better (0.266).
DeepSets using phenotype qualifiers neither consistently outperformed shuffled-metadata baselines nor generalized to unseen diseases.
On 230 ablated cases, state-safe scoring reached only 0.050 MRR (versus 0.066 naive, 0.050 omission). These remedies proved insufficient, though untested downstream architectures may perform differently.

\section{Discussion}

Our study focuses on rare diseases and retrieval-based diagnostic rankers rather than routine inpatient care, clinical outcomes, or the internal weights of foundation models. This scope is also a strength: rare-disease genetics provides a traceable link between published case reports and curated HPO annotations. Historical HPO releases allow us to observe when source-linked evidence appears, although only at release intervals rather than exact dates.

These results demonstrate that high diagnostic accuracy is at times achieved through access to the published case report describing the patient. When we kept the patient case and ontology release identical, removing clues linked to that publication caused diagnostic performance to plunge.
Background HPO growth, removing evidence for competing diseases, and adding synthetic symptoms did not reproduce this effect.
Removing evidence linked to a case report can also remove facts supported by other sources. The test shows reliance on that report, not an inability to reason.

Medical diagnosis is fundamentally an act of prospective inference under uncertainty, not retrospective retrieval. A benchmark cannot claim to measure clinical reasoning if an algorithm has covert access to the very publication that solved the case. Meaningful evaluation therefore rests on three distinct epistemic boundaries: \textbf{patient separation} isolates the individual from training; \textbf{evidence separation} removes their case reports from reference knowledge; and \textbf{time-correct evaluation} bounds reference data to what was known at the point of care. Unrestricted modern knowledge remains vital for assessing an AI's operational utility as a contemporary bedside assistant. However, claiming genuine diagnostic generalization demands complete evidence independence.
Provenance-aware reporting (Appendix~\ref{app:reporting}) provides a practical path toward benchmarks that distinguish dependence on published cases from diagnostic generalization.

\acks{This work was supported by the Innovative Medicines Initiative 2 Joint Undertaking (JU) under grant agreement No.~101034427 (Screen4Care). The JU receives support from the European Union's Horizon 2020 research and innovation programme and EFPIA. L.G.P. acknowledges support from the MIST project, funded by the Novo Nordisk Foundation under grant reference number NNF25OC0103204. Part of the computation for this project was performed on the UCloud interactive HPC system managed by the eScience Center at the University of Southern Denmark.}

\clearpage
\bibliography{references}

\clearpage
\appendix
\section{Additional Analyses and Interpretation}
\label{app:findings-expanded}

\subsection{Alternative Explanations Tested}
\label{app:controls}
The main analysis asks whether a case becomes easier to diagnose when evidence
from its source publication enters HPO. We tested five other explanations:

\begin{itemize}
    \item \textbf{Was every case improving over time?}
    We compared publications at the ingestion event with publications from the
    same period that had no event.

    \item \textbf{Did HPO's structure or disease list change?}
    We repeated the analysis while holding the HPO hierarchy and candidate diseases
    fixed.

    \item \textbf{Did removal affect the correct disease specifically?}
    We removed source-linked annotations for the target disease separately from
    annotations for competing diseases.

    \item \textbf{Would any new phenotypes produce the same gain?}
    We compared the real additions with synthetic additions matched on number,
    information content, ontology depth, and organ system.

    \item \textbf{Was the result driven by shared citations or a single release?}
    We repeated the analysis using only annotations cited exclusively to the
    test publication, excluding the large 2024 ingestion batch, requiring
    annotations to persist, and retaining one copy of each repeated profile.
\end{itemize}

\subsection{What the Controls Showed}
\label{app:control-results}
The effect was specific to source-linked evidence about the correct disease
(Table~\ref{tab:findings-results}). Publications without an ingestion event
changed by only +0.002. The event-versus-control difference was +0.580.
Removing target-disease evidence reproduced the main interaction (+0.792);
removing evidence for competing diseases produced almost no change (+0.001).
Real phenotype additions also outperformed matched synthetic additions by
+0.772.

The effect remained positive when we removed only annotations citing the test
publication and no other source (+0.372; 95\% CI 0.303--0.442). This estimate
uses the exclusive-source subset of 868 cases from 123 publications. Because
both the removal policy and cohort differ from the primary analysis, the
smaller effect cannot be attributed to the removal policy alone.
The result also remained consistent when we fixed the HPO hierarchy and candidate
diseases, required persistent annotations, removed duplicate profiles, or
excluded the largest ingestion batch.

In a separate historical comparison, cases whose publications were later
ingested improved by +0.640; cases without an observed ingestion event changed
by +0.004. These results identify source-linked evidence as the leading
explanation in this cohort. They do not show that every retrospective benchmark
has this problem or reveal which patient in a multi-patient paper produced an
annotation.

\subsection{What the Study Does Not Establish}
Several limits matter:

\begin{itemize}
    \item \textbf{The exact ingestion date is unknown.}
    We observe snapshots of HPO, not every change between releases. The event is
    therefore the first release in which a citation appears, not necessarily the
    day it was added.

    \item \textbf{A publication citation is not a patient-level link.}
    It identifies the source document, but not the exact patient, sentence, or
    curator decision. Removing an annotation with several citations can also
    remove evidence supported by another publication.

    \item \textbf{The cohort was selected for observable ingestion events.}
    The effect size describes this cohort, not the prevalence or average size of
    source dependence across all clinical benchmarks.

    \item \textbf{The evidence-loss audits concern recorded information.}
    They do not determine biological disease onset or prove that a missing
    annotation is a curation error.

    \item \textbf{The mitigation results apply only to the methods tested.}
    They do not prove that every possible downstream model must fail.

    \item \textbf{Reciprocal rank is not a clinical outcome.}
    A change in rank shows an evaluation dependency, not improved safety,
    diagnostic decisions, or patient benefit.
\end{itemize}

\subsection{Reporting Checklist}
\label{app:reporting}
For literature-derived clinical benchmarks, authors should report:

\begin{enumerate}
    \item \textbf{Where test cases came from.}
    Preserve publication identifiers and report any missing provenance.

    \item \textbf{What knowledge the system could access.}
    Give knowledge-resource and ontology versions, retrieval dates, and external resources
    available during prediction.

    \item \textbf{Whether test sources overlap with prediction knowledge.}
    Report this separately from patient-level deduplication and explain how
    multiply cited assertions are handled.

    \item \textbf{Which cases could actually be evaluated.}
    Report missing targets, mapping failures, empty profiles, and execution
    failures. State whether they remain in the denominator. In this study,
    unranked and failed cases receive reciprocal rank zero.

    \item \textbf{Whether the knowledge matches the intended time point.}
    Historical claims require historically available knowledge. This remains
    different from prospective clinical validation.
\end{enumerate}

These items help distinguish performance supported by independent evidence from
performance that depends on the publication used to create the test case.

\subsection{How Evidence Loss Was Tested}
\label{app:audits}
We ran two complementary audits. First, we tested loss of clinical order in
electronic records. Starting with 3,160 diagnosis-pair relationships from
MIMIC-IV \citep{johnson2023mimiciv}, we removed whole encounters from each
patient history. We compared this with removing whole patients while keeping
the remaining histories complete. Both strategies retained the same expected
amount of data, so their difference isolates the effect of fragmented histories.
We also reduced timestamp precision from minutes to days and weeks in MIMIC-IV
and eICU \citep{pollard2018eicu}.

Second, we tested how much published evidence survives curation. We followed
9,722 disease--phenotype pairs from Phenopacket Store cases
\citep{danis2025phenopacketstore} published through 2020. We checked whether
each pair appeared in HPO 2021 \citep{gargano2024hpo}, and then in PrimeKG
\citep{chandak2023primekg} or OpenBioLink \citep{breit2020openbiolink}.
These datasets represent different stages of the evidence pathway; they are not
one patient cohort.

\subsection{What the Evidence-Loss Audits Showed}
\label{app:evidence-loss-results}
With complete MIMIC-IV histories, the recorded order was observable for 76.8\%
of diagnosis pairs. After retaining only 25\% of encounters, this fell to
12.9\%, although both diagnoses were still present for 79.4\% of pairs.
When the same expected amount of data was removed by sampling whole patients
instead, 91.9\% of relationships remained detectable. Fragmentation therefore
lost more recoverable information than an equal reduction in data volume.

Coarser timestamps also erased order. At weekly resolution, 83.3\% of MIMIC-IV
and 90.0\% of eICU diagnosis-pair observations became ties. The dominant change
was loss of order, not reversal of order.

Of the 9,722 published disease--phenotype pairs, 3,454 (35.53\%) appeared as
positive HPO 2021 annotations. Capture was predictable in diseases excluded
from model fitting (AUC 0.715), indicating that selection was structured rather
than uniform. Of the pairs captured by HPO, 56.28\% appeared in PrimeKG or
OpenBioLink, compared with 4.95\% of pairs absent from HPO. Because these
resources partly reuse HPO, this is downstream propagation, not independent
confirmation. Failure to capture a pair does not by itself imply curation error.

\subsection{Primary Source-Reuse Result}
In the locked Resnik analysis, publication-weighted mean reciprocal rank rose
from about 0.26 before source ingestion to 0.84 at ingestion. Removing every
annotation citing the test publication from the ingestion release reduced it
to 0.05, without changing the patient query.

From before ingestion to ingestion, reciprocal rank changed by +0.582 with
standard knowledge and by $-0.212$ after source removal. Their difference was
+0.793 (95\% CI 0.751--0.835; Table~\ref{tab:findings-results}). Among 991
cases, 974 interactions were positive, none were negative, and 17 were tied.
IC-cosine and LIRICAL showed the same direction of source dependence. Because
the rankers use different candidate sets and scoring rules, their absolute
effect sizes should not be compared directly.

\subsection{What the Mitigation Experiments Showed}
\label{app:mitigation}
Adding recorded phenotype frequency outperformed shuffled frequency by +0.070
MRR, showing that frequency contained useful information. However, it did not
improve significantly over Resnik overall (+0.006 MRR; 95\% CI
$-0.008$ to $+0.019$): it helped on RAMEDIS but reduced performance on
Matchmaker Exchange and HMS.

The handling of negative findings also mattered. Treating missing disease
penetrance as certainty produced MRR 0.074 on temporally unseen cases. Applying
negative penalties only when penetrance was recorded raised MRR to 0.229, but
omitting negative evidence was still better at 0.266. Correcting the unsupported
default reduced the error; it did not make negative evidence consistently
helpful.

A DeepSets model using the preserved qualifiers improved over a flat
representation for later cases of previously seen diseases. It did not
consistently beat shuffled metadata and did not generalize to unseen diseases.
In the separate 230-case source-removal experiment, the prespecified
observation-aware policy achieved MRR 0.050, compared with 0.066 for naive
scoring and 0.050 when negative evidence was omitted. It failed all
prespecified success criteria. These results limit the tested approaches; they
do not rule out every possible model.

\subsection{How Failed or Uncovered Cases Were Counted}
\label{app:failures}
Source removal can leave a disease with no usable phenotype profile. Such cases,
missing targets, and execution failures remained in the assigned denominator
with reciprocal rank zero; they were not silently dropped.

For IC-cosine, 3,869 of 3,964 case-state rows had a ranked target; 95 target
profiles were empty. LIRICAL used a fixed 2026 ontology to map historical
identifiers. Unresolved terms were dropped and counted. Across 3,652 unique
tasks expanded to 3,964 case-state rows, 3,869 targets were ranked and 95 were
not. No run failed during execution.

\section{How the Diagnostic Rankers Work}
\label{app:rankers}

\subsection{Reciprocal Rank and the Ingestion Comparison}
\label{app:metrics}
Reciprocal rank gives more credit when the correct disease appears near the top
of a ranked list. If its rank is $r$, then $RR=1/r$. A correct disease ranked
first, second, or tenth therefore receives 1, 0.5, or 0.1. An unranked disease
receives zero, and tied diseases receive the reciprocal of their midpoint rank.

Mean reciprocal rank (MRR) averages these values. In the primary analysis, we
first average cases from the same publication, then give every publication
equal weight. Thus, 0.26, 0.84, and 0.05 are average reciprocal-rank values.
They are not percentages of correct diagnoses or the rank of one patient.

The source-ingestion effect compares two changes: how ranking changes when a
publication enters standard HPO, and how ranking changes over the same releases
after removing assertions citing that publication. A positive interaction means
that the standard knowledge gained more than the source-removed knowledge.
We resample whole publications, with all their cases together, when calculating
confidence intervals. This accounts for cases that share a source. The result
measures dependence on source-linked evidence; it does not by itself prove
model memorization or clinical benefit.

\subsection{Resnik: Matching Related Phenotype Terms}
\label{app:resnik}
Resnik similarity gives a high score when two phenotype terms share a specific
ancestor in the HPO hierarchy \citep{resnik1995information}. For example, two
different abnormalities can still match through a shared parent concept.
A rare, specific shared ancestor provides more information than a broad one.

The following equations state the exact implementation. Let $\mathcal D_K$ be
the candidate diseases in knowledge state $K$. The information content of
ancestor $a$ is higher when it appears in fewer disease profiles:
\begin{equation}
IC_K(a)=-\log\frac{df_K(a)+1}{|\mathcal D_K|+1},
\end{equation}
where $df_K(a)$ is the number of disease profiles containing $a$ after ancestor
propagation.

For each patient term $q$, the method finds the disease term $d$ with the most
informative common ancestor:
\begin{equation}
m_K(q,D)=\max_{d\in H_D}
IC_K\!\left(\operatorname{MICA}(q,d)\right),
\end{equation}
where $H_D$ is the positive phenotype profile of disease $D$. It then normalizes
each match by the information in the patient term and averages across mapped
patient terms $Q'$:
\begin{equation}
s_R(Q,D;K)=\frac{1}{|Q'|}
\sum_{q\in Q'}\frac{m_K(q,D)}{IC_K(q)}.
\end{equation}

This score is directional: it asks how well the disease explains the patient's
terms, without separately averaging disease-to-patient matches. Resnik's paper
provides the shared-information principle; the normalization and averaging are
specific to this implementation. Disease profiles and information-content
values are recalculated after source removal.

\subsection{IC-cosine: Comparing Whole Phenotype Profiles}
\label{app:cosine}
The IC-cosine ranker represents the patient and each disease as vectors over
HPO terms and their ancestors. Present terms receive their information-content
weight; absent terms receive zero. If $A_K(X)$ contains the terms in $X$ and
their ancestors, then
\begin{align}
v_K(X)_a&=IC_K(a)\,\mathbf 1\{a\in A_K(X)\},\\
s_C(Q,D;K)&=
\frac{v_K(Q)\cdot v_K(H_D)}
{\|v_K(Q)\|_2\,\|v_K(H_D)\|_2}.
\end{align}

In plain terms, the score is high when the patient's complete weighted profile
points in the same direction as the disease profile. Unlike Resnik
best-match averaging, every weighted term contributes to one whole-profile
comparison. All vectors and weights are rebuilt after source removal. If source
removal leaves the target disease with an empty profile, that case receives
reciprocal rank zero (Appendix~\ref{app:failures}).

\subsection{LIRICAL: Combining Clinical Likelihood Ratios}
\label{app:lirical}
LIRICAL asks how much each finding favors one disease over alternatives
\citep{robinson2020lirical}. Conceptually, for finding $h$ and disease $D$,
\begin{equation}
LR(h,D)=\frac{P(h\mid D)}{P(h\mid\neg D)}.
\end{equation}
A finding that is common in $D$ but uncommon in other diseases provides stronger
support for $D$. LIRICAL combines this evidence across findings; its original
paper describes the complete algorithm and ontology matching.

We ran LIRICAL 2.4.1 in phenotype-only, global-prioritization mode. The patient
queries contained observed phenotypes only; no genotype evidence was added.
Standard and source-removed runs used the corresponding historical HPO
annotations. A fixed 2026 ontology mapped historical identifiers, unresolved
terms were dropped and counted, and ORPHA targets used
\texttt{--use-orphanet}. Because LIRICAL and the semantic rankers use different
candidate sets and scoring rules, we compare the direction of the paired
source-removal effect, not their absolute scores. Failure counts appear in
Appendix~\ref{app:failures}.

\subsection{Methods Tested for Handling Incomplete Observations}
\label{app:mitigation-methods}
These supporting experiments ask whether additional modeling can recover useful
information from incomplete clinical records. They are separate from the main
source-removal test.

\textbf{Frequency-aware scoring.}
On 752 external RareBench cases \citep{chen2024rarebench}, recorded phenotype
frequency adjusted the semantic-match score. A less frequent disease phenotype
could contribute less; missing frequency was treated as neutral. Separate tests
of negative findings compared three choices: treating missing penetrance as
certainty, applying a penalty only when penetrance was recorded, or ignoring
negative findings.

\textbf{DeepSets residual model.}
Deep Sets can process an unordered collection of phenotype observations
\citep{zaheer2017deepsets}. Here it learned a small correction to the IC-cosine
score:
\begin{equation}
s_\theta(Q,D)=s_C(Q,D)+0.10\tanh g_\theta(Q,D).
\end{equation}
The correction cannot exceed 0.10 in either direction. Training used the
pairwise loss
\begin{equation}
\mathcal L=\mathbb E\,
\operatorname{softplus}\{-10(s_\theta^+-s_\theta^-)\},
\end{equation}
which rewards a higher score for the target disease than for a comparison
disease. Flat, shuffled, and random metadata tested whether gains came from
meaningful qualifiers rather than added model capacity. Publication, disease,
and temporal splits tested different forms of generalization. These neural
results are exploratory.

\textbf{Locked source-removal experiment.}
A separate locked cohort of 230 cases tested observation-aware scoring after
source removal. Success required improvement over naive scoring,
noninferiority to omitting negative evidence within 0.01 MRR, and improvement
over shuffled exclusions. Results appear in Appendix~\ref{app:mitigation}.
These criteria evaluate the selected methods, not every possible algorithm.

\end{document}